\documentclass[journal,11pt]{IEEEtran}

\usepackage{cite}

\ifCLASSINFOpdf
   \usepackage[pdftex]{graphicx}
\else
\fi

\usepackage{amsmath}

\usepackage{wrapfig}

\usepackage{graphicx}
\usepackage{subcaption}
\usepackage[latin1]{inputenc}
\usepackage{amssymb}
\usepackage{multirow}
\usepackage{rotating}
\usepackage{floatflt}

\usepackage{algorithm}
\usepackage{algpseudocode}

\usepackage{bm}
\usepackage{xspace}
\usepackage{amssymb}
\usepackage{amsmath}
\usepackage{amsfonts}
\usepackage{euscript}

\newcommand{\mF}{{\mathcal F}}
\newcommand{\mI}{{\mathcal I}}

\newcommand{\mL}{{\mathcal L}}

\newcommand{\mS}{{\mathcal S}}

\newcommand{\mM}{{\mathcal M}}

\newcommand{\mbbR}{{\mathbb R}}

\def\registered{{\ooalign {\hfil\raise .05ex\hbox{\scriptsize
R}\hfil\crcr\mathhexbox20D}}}

\def\REgistered{{\ooalign
{\hfil\raise.09ex\hbox{\tiny \sf R}\hfil\crcr\mathhexbox20D}}}

\makeatletter
\DeclareRobustCommand\onedot{\futurelet\@let@token\@onedot}
\def\@onedot{\ifx\@let@token.\else.\null\fi\xspace}

\graphicspath{{figsFinal/}}

\DeclareMathOperator*{\argmax}{arg\,max}
\DeclareMathOperator*{\argmin}{arg\,min}

\usepackage{booktabs}
\usepackage{color}
\newcommand{\revision}[1]{#1}

\begin{document}
%
\title{Keypoint Transfer for \\Fast Whole-Body Segmentation}

\author{Christian Wachinger, Matthew Toews, Georg Langs, William Wells, Polina Golland
\thanks{C. Wachinger was with  Computer  Science and Artificial Intelligence Lab (CSAIL) at the Massachusetts Institute of Technology (MIT) and the Department of Neurology, Massachusetts General Hospital, Harvard Medical School when most of the work was performed. He is now with the lab for Artificial Intelligence in Medical Imaging (AI-Med), Child and Adolescent Psychiatry, LMU München.}
\thanks{ 
P. Golland, G. Langs, and W. Wells are with CSAIL at MIT.   
M. Toews is with the  Ecole de Technologie Superieure, Montreal. 
G. Langs is with the Computational Imaging Research Lab, Medical University of Vienna. 
W. Wells is with the Brigham and Women's Hospital, Harvard Medical School. 
}
}

\markboth{Transactions on Medical Imaging}%
{Wachinger \MakeLowercase{\textit{et al.}}: Keypoint Transfer Segmentation}

\maketitle

\begin{abstract}
We introduce an approach for image segmentation based on sparse correspondences between keypoints in testing and training images. 
Keypoints represent automatically identified distinctive image locations, where each keypoint correspondence suggests a transformation between images. 
We use these correspondences to transfer label maps of entire organs from the training images to the test image.
The keypoint transfer algorithm includes three steps: (i) keypoint matching, (ii) voting-based keypoint labeling, and (iii) keypoint-based probabilistic transfer of organ segmentations.  
We report segmentation results for abdominal organs in whole-body CT and MRI, as well as in contrast-enhanced CT and MRI.  
Our method offers a speed-up of about three orders of magnitude in comparison to common multi-atlas segmentation, while achieving an accuracy that compares favorably. 
Moreover, keypoint transfer does not require the registration to an atlas or a training phase. 
Finally, the method allows for the segmentation of scans with highly variable field-of-view. 
\end{abstract}



\IEEEpeerreviewmaketitle


\section{Introduction}
Is atlas-based segmentation without dense correspondences possible? 
Dense correspondences, i.e., correspondences for each location in the test image to the training images, are computed by 
common registration- and patch-based segmentation methods~\cite{Coupe2011,rousseauHS11,heckemann2006automatic,rohlfing2004evaluation,langerak2010label,sabuncu2010generative,wachinger2015contour,wachinger2017efficient}. 
The computation of dense correspondences via deformation fields or the identification of similar patches can become computationally intense for images with a large field-of-view.
As an alternative, we introduce an approach for image segmentation based on sparse correspondences by identifying distinctive locations in the image:  \emph{keypoints}. 
Keypoints are automatically computed as local optima of a saliency function~\cite{lowe}, contrary to manually selected landmarks~\cite{potesil2014learning}. 
We match keypoints between test and training images to establish correspondences for a sparse set of image locations. 
Based on these correspondences, the segmentation masks of entire organs are transferred and fed into a probabilistic fusion algorithm.  
The segmentation accuracy compares favorably to common multi-atlas techniques, while working with sparse correspondences leads to a  computationally efficient algorithm, offering orders of magnitude of speed-up.

%
\begin{figure*}[t]
\begin{center}
\includegraphics[width=1.0\textwidth]{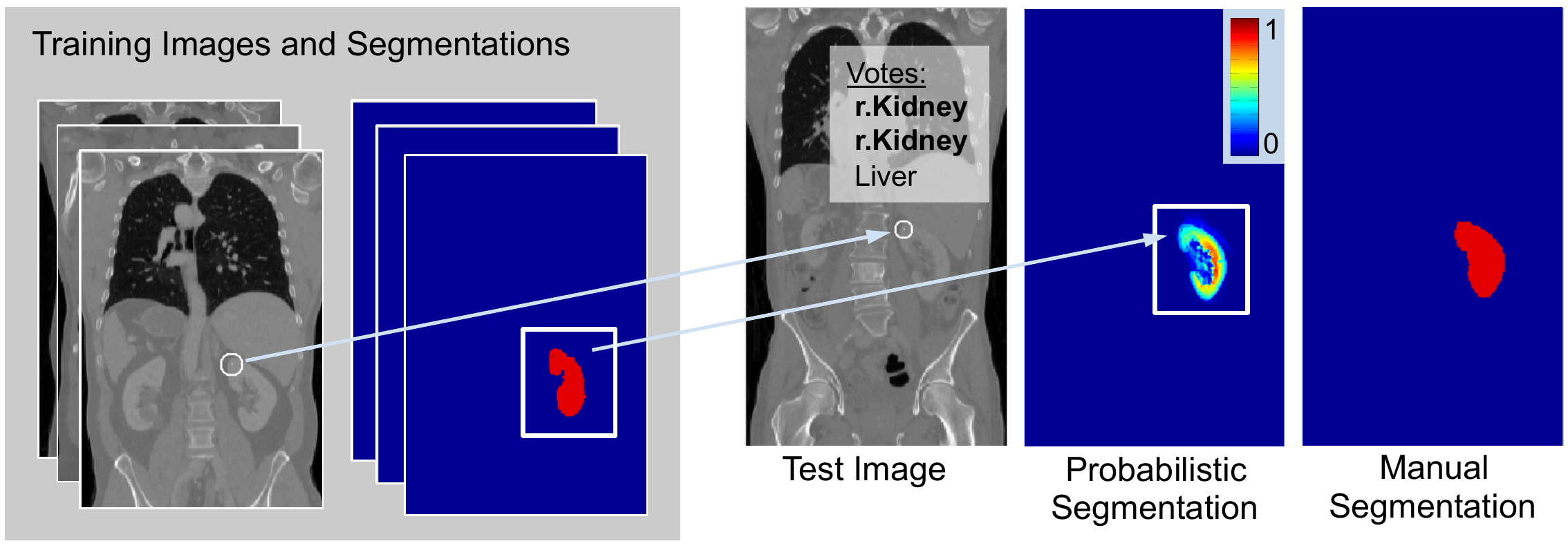} 
\caption{Overview of the keypoint transfer algorithm for whole-body segmentation. First, keypoints (white circles) are matched between training and test images (arrow). Second, the organ label of the test keypoint is voted on, based on the identified matches (r.Kidney). Third, training keypoints with r.Kidney as label transfer the surrounding organ map to the test image, creating a probabilistic segmentation. The manual segmentation is shown for comparison. %
 \label{fig:am}}
\end{center}
\end{figure*}

We outline the keypoint transfer segmentation algorithm in Fig.~\ref{fig:am} \revision{and as animation in the supplementary material}. 
Keypoints are extracted at salient image regions and described by their geometry and a descriptor based on a histogram of local image intensity gradients.
Following keypoint extraction, we segment an image in three steps.
First, we match keypoints in the test image to keypoints in the training images based on the geometry and the descriptor.  
Second, keypoint labels are voted on based on matches. 
In the example in Fig.~\ref{fig:am}, the keypoint receives two votes for right kidney and one for liver, resulting in a majority vote for right kidney. 
Third,  the label mask is transferred for the entire organ for each match that is consistent with the majority label vote. 
The organ map from one training image is possibly transferred multiple times if more than one match is available for this training image. 
Keypoint transfer also integrates the certainty in the keypoint label voting and computes the intensity similarity between scans. 
The algorithm's capability in approximating the organ shape  can further improve with a growing number of manually labeled scans, where additional images can be included in the training set without the need for a dedicated training stage. 

In addition to being fast, keypoint transfer is beneficial for segmenting images with varying field-of-view. 
In our experiments, we use manually annotated whole-body scans to segment images with a limited field-of-view. 
Such scans are commonly acquired in clinical practice by focusing on the region of diagnostic interest, and thereby reducing scanning time and radiation dose. 
The intensity-based registration between  images with a limited field-of-view and full abdominal images is challenging, particularly when anatomical structures are initially not approximately aligned.
Keypoint matching is robust to such variations in field-of-view and therefore offers an efficient and practical tool to deal with the growing number of clinical images.
A preliminary version of this work was presented at a conference~\cite{wachinger2015keypoint}. 
\revision{Major changes in this version are: an updated presentation of the algorithm in the method's section including an algorithmic perspective of the keypoint segmentation, the addition of experiments for whole-body MR and contrast-enhanced MR on the gold corpus, an evaluation of the matching criteria on segmentation performance, and additional experiments on all four contrasts on the silver corpus.  }

\subsection{Related Work}
For the segmentation of large field-of-view scans, several methods have previously been proposed. 
A combination of discriminative and generative models~\cite{iglesias2011combining} and entangled decision forests~\cite{montillo2011entangled}   have been explored for the segmentation of CT scans. 
Authors in~\cite{lay2013rapid} proposed simultaneous segmentation of multiple organs by combining global and local context. 
Marginal space learning was described for organ detection in~\cite{zheng2009marginal}. 
\revision{A combination of multi-object recognition and iterative graph-cuts with  active shape model was introduced in~\cite{chen20113d}.}
Random decision forests for patch-based segmentation were proposed in~\cite{konukoglu2013neighbourhood}.
Contrary to previously demonstrated methods, keypoint transfer does not need an extensive training on manually annotated images. 
%


We use the publicly available Visceral dataset~\cite{langs2013visceral,jimenez2016cloud} for evaluating keypoint transfer. 
%
Methods based on multi-atlas segmentation have been applied on the Visceral data~\cite{gokselsegmentation,delhierarchical,gass2014multi,heinrich2015multi,kechichian2014automatic,kahl2015good}, which we employ  as a baseline  in our evaluation. 
Multi-atlas segmentation with atlas selection and label fusion was proposed for 12 abdominal structures on clinically acquired CT in~\cite{xu2015efficient}. 
We use a 3D extension~\cite{toews2013efficient}  of the popular scale invariant feature transform~(SIFT)~\cite{lowe} for the extraction and description of keypoints. 
Next to image registration, 3D SIFT features were also used for studying questions related to neuroimaging~\cite{toews2010feature} and for efficient big data analyses of medical images with approximate nearest-neighbor search~\cite{toews2015feature}. 
Contrary to previous applications of the 3D SIFT descriptor, we use it to propagate information across images.

\section{Method}
Given training images\footnote{\revision{We use the term "training images" although our algorithm does not have an explicit training stage.}} $\mI = \{I_1, \ldots, I_n\}$ and corresponding segmentations $\mS = \{S_1, \ldots, S_n\}$, where $S_i(x) \in \{1, \ldots, \eta\}$ for $\eta$  labels,  
our aim is to infer segmentation~$S$ for test image~$I$. 
To this end, we automatically identify keypoints in the images and employ them to create sparse correspondences. 
This is in contrast to atlas-based segmentation, where images are aligned with deformable registration. 
%
Keypoints are extracted by locally maximizing a saliency function.  
Following the SIFT descriptor, we use the difference-of-Gaussians~\cite{lowe}
\begin{equation}
\{(x_i, \sigma_i)\} = \text{local} \, \argmax_{x,\sigma} | f(x, \kappa \sigma) - f(x, \sigma) |,
\end{equation}
where~$x_i$ and $\sigma_i$ are the location and scale of keypoint~$i$, $f(\cdot,\sigma)$ is the convolution of the image $I$ with a Gaussian kernel of variance~$\sigma^2$, and $\kappa$ is a multiplicative scale sampling rate. 
Keypoints are located at distinctive spherical image regions that show a local extrema in scale-space.
For the descriptor of the keypoint~$F^D$, we employ  a 3D extension of the image gradient orientation histogram~\cite{toews2013efficient} with 8 orientation and 8 spatial bins, which is scale and rotation invariant and further robust to small deformations. 
Working with image gradients instead of intensity values makes the descriptor more robust to intensity variations and therefore offers advantages in comparing descriptors across subjects. 

%
The keypoint~$F$ of a salient image region is described by the 64-dimensional histogram~$F^D$ with the \revision{keypoint location}~$F^x \in \mbbR^3$ and \revision{keypoint scale}~$F^\sigma \in \mbbR$, resulting in a compact 68-dimensional representation. 
%
Let~$F_I$ denote the set of keypoints identified in the test image~$I$ and let $\mF_\mI = \{\mF_{I_1}, \ldots, \mF_{I_n} \}$ denote the set of keypoints identified in the training images~$\mI$. \revision{Script letters are used for denoting training data and non-script letters for denoting testing data.} 
Each keypoint is assigned an organ label~$\mF_{I_i}$ according to the organ it is located in, $\mL = S_i(\mF^x)$ for $\mF \in \mF_{I_i}$. 
Keypoints that are located in the un-segmented background, are discarded. 
%
For the keypoints in the test image, the organ label~$L$ is unknown and inferred with a voting algorithm as described later in this section.   
\revision{Table~\ref{tab:notation} summarizes the notation used in the article.}

\subsection{Keypoint Matching \label{sec:matching}} 
In the first step, we match each keypoint in the test image with keypoints in the training images. 
To ensure high quality matches, a two-stage matching procedure is proposed to improve the reliability of the matches by including additional constraints. 
In the first stage, a match~$\mM(F)_i$ is computed between a test keypoint~$F \in F_I$ and keypoints in a training image $\mF_{I_i}$. To this end, we find the nearest neighbor based on the similarity of the descriptor and scale constraints
\begin{equation}
\mM(F)_i = \argmin_{\mF \in \mF_{I_i}} \| F^D - \mF^D \|,  \ \ \ \text{s.t.} \ \   \varepsilon_\sigma^{-1} \leq \frac{F^\sigma}{\mF^\sigma} \leq \varepsilon_\sigma,
\end{equation}
where a loose threshold on the scale allows for variations up to a factor of $\varepsilon_\sigma = 2$. 
The distance ratio test is employed to discard keypoint matches that are not reliable~\cite{lowe}, where we compute the ratio between  descriptors of the closest and second-closest neighbor. All matches with a distance ratio of greater than 0.9 are rejected. 

In the second stage, we improve the matches by additionally imposing loose spatial constraints, which requires an approximate alignment. 
For our datasets, accounting for translation was sufficient at this stage due to consistent patient orientation; as an alternative we could use an efficient keypoint-based pre-alignment~\cite{toews2013efficient}.
\revision{The most likely translation~$t_i$ suggested by the matches~$\mM_i$ is computed with the Hough transform~\cite{ballard1981ght}\footnote{\revision{As an alternative to the Hough transform, the within-sample median translation could be used.}}.}  
Mapping the training keypoints with $t_i$ leads to an approximate alignment of the keypoints and allows for an updated set of matches with an additional spatial constraint 
\begin{align}
&\mM(F)_i = \argmin_{\mF \in \mF_{I_i}} \| F^D - \mF^D \|,  \\
& \text{s.t.} \ \  \varepsilon_\sigma^{-1} \leq \frac{F^\sigma}{\mF^\sigma} \leq \varepsilon_\sigma, \ \ \| \revision{F^x - \mF^{x} - t_i} \|_2 < \varepsilon_x, \nonumber
\end{align}
where a spatial threshold $\varepsilon_x$ is selected to keep 10\% of the closest matches. 
As previously, matches that do not fulfill the distance ratio test are discarded. 
\revision{Note that the estimated global translation between training and testing image is only used for improving matches.}

\begin{table}
\revision{
  \begin{tabular}{ll}
  \toprule
  $I, \mI = \{I_1, \ldots, I_n\}$ & Test and training images \\
  $S, \mS = \{S_1, \ldots, S_n\}$ & Test and training segmentations \\
  $S^l$ & Segmentation probability map for label $l$ \\
  $\hat{S}$ & Inferred segmentation \\
  $F, \mF$ & Keypoint in test and training image \\
  $F^x, \mF^x$ & Location of test and training keypoint \\
  $F^\sigma, \mF^\sigma$ & Scale of test and training keypoint \\
  $F^D, \mF^D$ & Descriptor of test and training keypoint \\
  $F_I, \mF_{I_i}$ & All keypoints from test and training image $i$ \\
  $L, \mL$ & Label of test and training keypoint \\
  $\hat{L}$ & Inferred label for test keypoint\\
  $m$ & Match between a test and training keypoint \\
  $m(F)$ & Corresponding training keypoint matched to $F$ \\
  $\mM(F)$ & All matches for test keypoint $F$ \\ 
  $\mM_i$ & All matches for training image $I_i$ \\
  $\mI_m, \mS_m$ & Training image and segmentation of match $m$ \\
  \bottomrule
  \end{tabular}
  \caption{\revision{Summary of notation used in the article. }
\label{tab:notation}}}
 \end{table}

\begin{figure}
\centering
\includegraphics[width=0.45\textwidth]{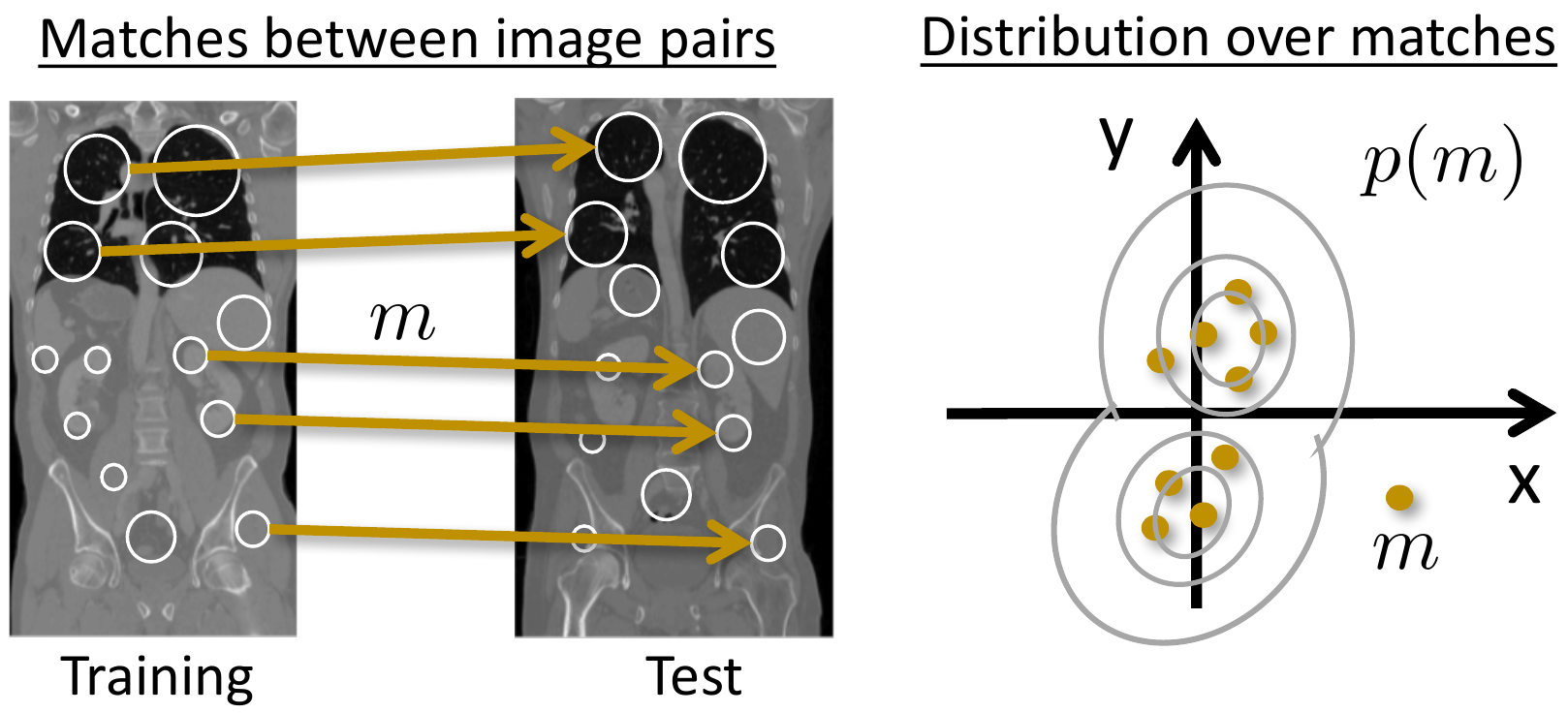}
\caption{\revision{Illustration of the construction of the distribution $p(m)$ over matches $m$. Arrows indicate matches between training and test image. Each match proposes a translation, illustrated as dots in 2D. Based on these samples the distribution is estimated.}
 \label{fig:distMatch}}
\end{figure}


At this stage, each keypoint $F$ in the test image is matched to at most one keypoint $\mM(F)_i$ per training image $I_i$, denoted as match $m$. 
\revision{We quantify the reliability of a match $m$ by constructing a distribution over matches $p(m)$, see Fig.~\ref{fig:distMatch}. 
We build the distribution based on the translation of a match, i.e., matches that propose a translation that is not proposed by other matches receive a low probability.
Hence, the probability~$p(m)$ for  a match~$m$ defines the translational consistency of the match with respect to  other matches.
This translational consistency only holds for a training and test image pair, so that a separate distribution is estimated for each training image. 
For the estimation, we use matches~$\mM_i$ between keypoints in the test image and those in the $i$-th training image and kernel density estimation. 
For notational ease, we write $p(m)$, although there are actually separate distributions $p^i(m)$ for each training image $i$;  
the selection of the corresponding distribution is evident from the training keypoint involved in the match.}
%
The non-parametric model accommodates  multi-modal distributions, which is  helpful for whole-body scans, as keypoints in the upper abdomen may suggest a different transformation than those in the lower abdomen.

\subsection{Keypoint Voting \label{sec:voting}}

\begin{figure}
\centering
\includegraphics[width=0.13\textwidth]{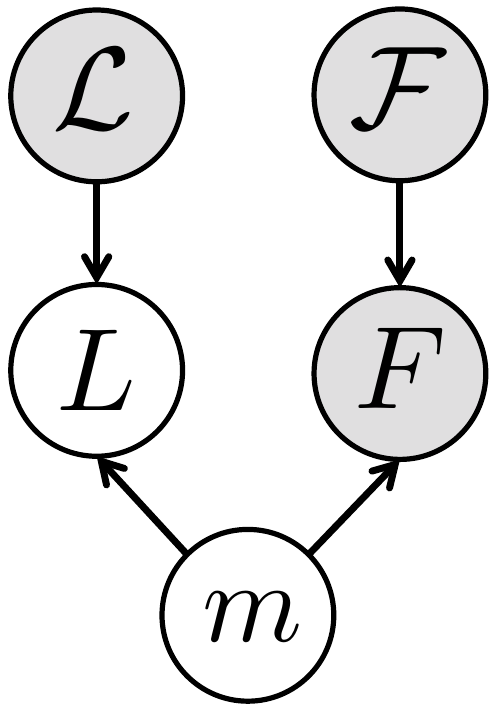}
\caption{Graphical model for keypoint voting with the match $m$ and the keypoint label $L$ being latent random variables. 
The labels of training keypoints $\mL$, the test keypoint $F$ and the training keypoints $\mF$ are observed, illustrated as shaded nodes.
 \label{fig:probModel}}
\end{figure}

After computing matches for keypoints in the test image, an organ label~$L$ is inferred for each keypoint in the test image based on the generative model illustrated in Fig.~\ref{fig:probModel}. %
The latent variable~$m$ represents keypoint matches from the previous step. 
With keypoint labeling it is possible to get a coarse localization of organs in the image. %
Further, keypoint labels are employed to guide the image segmentation as described  in the following section. 
For inferring keypoint labels, we use the factorization from the graphical model in Fig.~\ref{fig:probModel} and marginalize over the latent random variable~$m$ 
\begin{align}
p(L, F, \mL, \mF ) &= \sum_{m \in \mM(F)} p( L , F, \mL, \mF, m)  \\ 
&=   \sum_{m \in \mM(F)} p( L  | \mL, m) \cdot p(F | \mF, m) \cdot p(m ),  \label{equ:featLike}
\end{align}
where $\mM(F)$ includes matches for keypoint~$F$ across all training images. 
Working with a sparse set of matches makes the marginalization computationally efficient. 
The label probability is defined as
\begin{align}
p(L = l | \mL, m ) = \left\{ \begin{array}{rl}
 1 &\mbox{ if $\mL_{m(F)} = l$}, \\
  0 &\mbox{ otherwise,}
       \end{array} \right.
\end{align}
where $\mL_{m(F)}$ is the label of a training keypoint that the match~$m$ assigns to the test keypoint~$F$. 
Based on the descriptor, the keypoint probability is defined as
\begin{equation}
p(F | \mF, m) = \frac{1}{\sqrt{2\pi \tau^2}} \exp \left( -\frac{ \| F^D - \mF^D_{m(F)} \|_2^2}{2\tau^2} \right), \label{equ:keyLike}
\end{equation}
where we set $\tau^2 =  \max_m \| F^D - \mF^D_{m(F)} \|_2^2$. 
\revision{The most likely organ label is assigned to be the label of  the keypoint~$\hat{L}$}
\begin{align}
\hat{L} &= \argmax_{l \in \{1, \ldots, \eta \} }  p(L = l | F, \mL, \mF) \\
&= \argmax_{l \in \{1, \ldots, \eta \} }  p(L = l,  F, \mL, \mF). \label{equ:featVote}
\end{align}
\revision{To summarize, each training keypoint $\mF$ that was matched to test keypoint $F$ votes for the label $L$ with the organ it is located in $\mL$. If we would set the probabilities $p(F | \mF, m) \propto 1$ and $p(m ) \propto 1$ to constant values this would result in a majority vote for the keypoint label. To increase robustness, we weight the contribution of each training keypoint by the similarity of the descriptors $p(F | \mF, m)$ and the probability of the match $p(m)$.}

\subsection{Keypoint segmentation}


\begin{algorithm}[t]
\caption{\revision{Keypoint transfer segmentation. Organ segmentations from the training images are transferred to the test image via identified matches.}\label{alg:kts}}
\begin{algorithmic}[1]
\State $\forall l \in \{1,\ldots,\eta\}, S^l = {\bf 0}$
\ForAll{$F \in F_I$}
	\ForAll{$m \in \mM(F)$}
		\If{$\hat{L} \neq \mL_m$} continue
		\EndIf
		\State W = {\bf 0}
		\ForAll{$x: \mS_m(x) = \hat{L}$}
			\State $W(x) = \frac{1}{ \sqrt{2\pi} \nu} \exp \left( - \frac{ (I(x) - \mI_m(x))^2}{ 2\nu^{2}} \right)$
		\EndFor
	\State $S^{\hat{L}} = S^{\hat{L}} + W * p(L = \hat{L}) * p(m)$		
	\EndFor
\EndFor
\State $\forall x: \hat{S}(x) = \argmax_{l \in \{1,\ldots,\eta\}} S^l(x)$
\State \Return $\hat{S}$
\end{algorithmic}
\end{algorithm}

\begin{figure}
\centering
\includegraphics[width=0.35\textwidth]{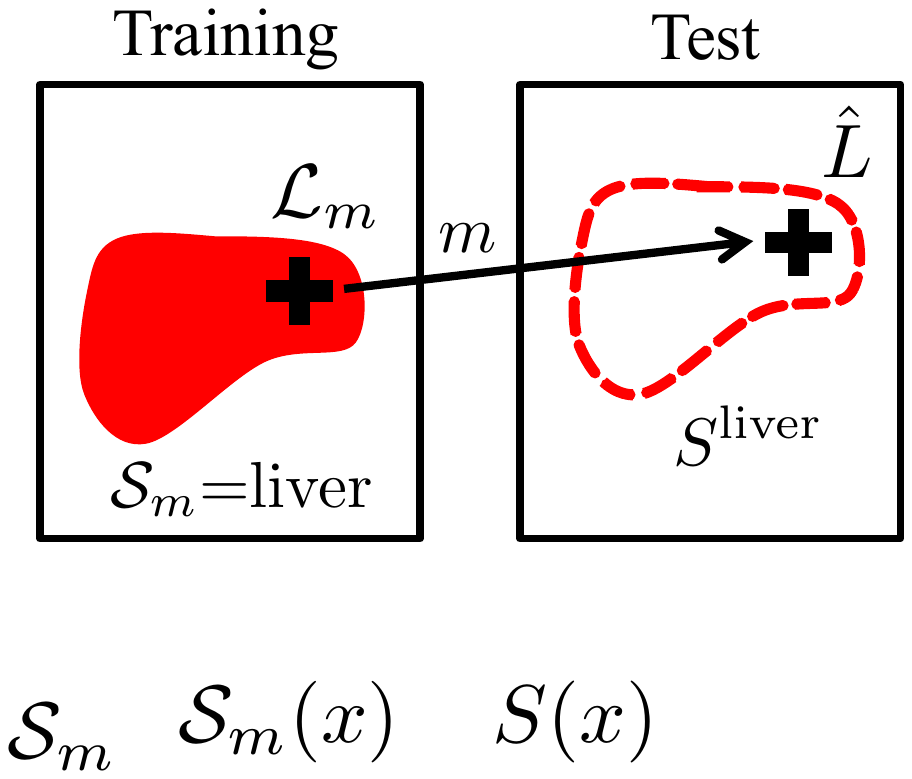}
\caption{Illustration of keypoint transfer segmentation for the example of liver. 
The crosses indicate keypoints in the training and test images with the match $m$, illustrated as arrow. 
For the label transfer to take place, the label of the training keypoint~$\mL_m$ and the voted label of the test keypoint~$\hat{L}$ have to be liver. 
The segmentation map of the training image $\mS_m = \text{liver}$ is then transferred and the probability map for liver $S^{\text{liver}}$ is updated. 
To increase the robustness and accuracy of the segmentation, we weigh the transferred segmentation according to the certainty in (i) the keypoint label voting, (ii) the match, and (iii) the local intensity similarity of the test and training image. 
 \label{fig:liver}}
\end{figure}

The keypoint segmentation is based on keypoint matching and voting from the previous sections. 
In~\cite{wachinger2015keypoint}, we derived the segmentation method from a generative model with marginalization over latent random variables. 
Here, we provide an algorithmic perspective in Algorithm~\ref{alg:kts} instead, to emphasize the simplicity of transferring entire organs maps. 
The method uses the extracted keypoints~$F$, the identified matches~$m$, and the voted label of the keypoint~$\hat{L}$, as presented in previous sections. 
The objective is to estimate the segmentation $\hat{S}$ based on probability maps for each of the organs $S^l$ with $l \in \{1,\ldots,\eta\}$. 
In short, the maps are computed by going through all the matches and transferring entire organ label maps from the training images, where the transformations are implied by the matches. 
Fig.~\ref{fig:liver} illustrates the procedure.

\begin{table*}[t]
 \centering
 \resizebox{\textwidth}{!}{
  \begin{tabular}{ccccccccccccc}
  \toprule
& Organs & Liver \ & Spleen \  & Aorta \ & Trachea \  & r.Lung  \ & l.Lung  \ & r.Kid \  & l.Kid \ & r.PM \ & l.PM \  & Bckgrnd  \\ 
\midrule
\parbox[t]{2mm}{\multirow{3}{*}{\rotatebox[origin=c]{90}{ceCT}}} & \# Keypts &13.6 & 4.0 & 7.6 & 3.0 & 29.7 & 24.7 & 12.1 & 12.2 & 2.5 & 3.0 & 526.0 \\ 
& \% Labeled & 73 & 89 & 98 & 100 & 95 & 92 & 98 & 99 & 94 & 92 & 33 \\ 
& \% Correct & 87 & 91 & 97 & 99 & 100 & 100 & 98 & 100 & 99 & 93 & 0 \\ 
\midrule
\parbox[t]{2mm}{\multirow{3}{*}{\rotatebox[origin=c]{90}{wbCT}}} &\# Keypts & 6.0 & 2.6 & 5.6 & 4.4 & 28.2 & 24.0 & 6.7 & 9.0 & 2.5 & 2.5 & 637.2 \\ 
&\% Labeled & 93 & 98 & 100 & 100 & 98 & 98 & 98 & 99 & 98 & 100 & 35 \\ 
&\% Correct & 82 & 87 & 92 & 100 & 99 & 99 & 98 & 96 & 100 & 93 & 0 \\ 
\midrule
\parbox[t]{2mm}{\multirow{3}{*}{\rotatebox[origin=c]{90}{ceMR}}} &\# Keypts & 25.1 & 4.0 & - & - & - & - & 9.5 & 11.8 & 2.5 & 1.8 & 562.9 \\ 
&\% Labeled & 92 & 95 & - & - & - & - & 99 & 100 & 91 & 82 & 59 \\ 
&\% Correct & 96 & 89 & - & - & - & - & 92 & 97 & 70 & 53 & 0 \\
\midrule
\parbox[t]{2mm}{\multirow{3}{*}{\rotatebox[origin=c]{90}{wbMR}}} &\# Keypts & 22.1 & 4.1 & 7.1 & 1.9 & 3.2 & 2.9 & 9.3 & 10.8 & 2.6 & 3.0 & 709.8 \\
&\% Labeled & 93 & 97 & 96 & 100 & 93 & 95 & 99 & 99 & 95 & 89 & 51 \\
&\% Correct & 93 & 83 & 49 & 99 & 87 & 87 & 90 & 93 & 75 & 65 & 0 \\
    \bottomrule
 \end{tabular}}
\caption{We report keypoint voting statistics per organ for ceCT, wbCT, ceMR, and wbMR: the average number of keypoints per organ, the average fraction of keypoints that get labeled, and the average fraction of correct keypoints labels. 
If there exists no reliable match, keypoints are not assigned labels . 
%
%
\label{tab:voting}}
\end{table*}

In details, we let $\mI_m$ denote the training image identified with match~$m$ after the transformation implied by the match has been applied. 
$\mS_m$ is similarly defined to be the selected and transformed segmentation map. 
We initialize the probability maps $S^l = 0$, and then iterate through all the keypoints in the test image and associated matches. 
We only allow for those training keypoints to transfer their segmentation whose votes are consistent with the majority vote in Eq.~(\ref{equ:featVote}), $\hat{L} = \mL_m$. 
Preventing keypoints with inconsistent label from transferring the segmentation improves the robustness of the algorithm. 
\revision{Instead of directly transferring the binary organ map, we modulate it with the local similarity between test and training image, $I(x) - \mI_m(x)$. 
Locations with similar intensities in test~$I$ and training~$\mI_m$ image obtain a higher weight than locations with larger intensity differences.} 
We use a Gaussian distribution for obtaining the weights~$W$ from the image differences, where~$\nu^2$ is the intensity noise variance. 
The probability map of the organ $\hat{L}$ is then incremented by the weights. 
In the increment, we also consider the certainty in the label voting and the match, by multiplying with the label probability $p(L = \hat{L})$ and the distribution over matches $p(m)$, respectively. 
Finally, we select the most likely label in the final segmentation~$\hat{S}$, where we account for not transferring the background surrounding the organ by assigning  the background label if the maximal probability in the voting is below 15\%. 

In the presented algorithm, keypoints can only transfer segmentations that have the same label, e.g., a liver keypoint can only transfer liver segmentation maps. 
This may be overly restrictive because the transformation resulting from a good match may also be usable for neighboring organs. 
\revision{Transferring the organ segmentation more often may lead to a better approximation of the target organ shape.}
In our experimental evaluation, we therefore also investigate the transfer of organ segmentations that are different from the keypoint labels  

We further study the potential benefit of accounting for affine organ variations across subjects. 
To this end, we estimate an organ-specific affine transformation when there are at least three matches for an organ between one training image and the test image. 
We use the random sample consensus~(RANSAC) algorithm~\cite{fischler1981random} to find the transformation parameters with the highest number of inliers. 
In our experiments, we have not observed a robust improvement of segmentation accuracy with the organ-wide affine transformation and therefore do not report it in the results. 
We believe that the reason for the affine transformation to not improve the results lies in the multiple transfer of organ labels per scan for  different translations; this, in combination with probabilistic weighting, already accounts for much of the organ variability. 

\begin{figure*}[t]
\begin{center}
    \begin{subfigure}[b]{0.49\textwidth}
        \includegraphics[width=\textwidth]{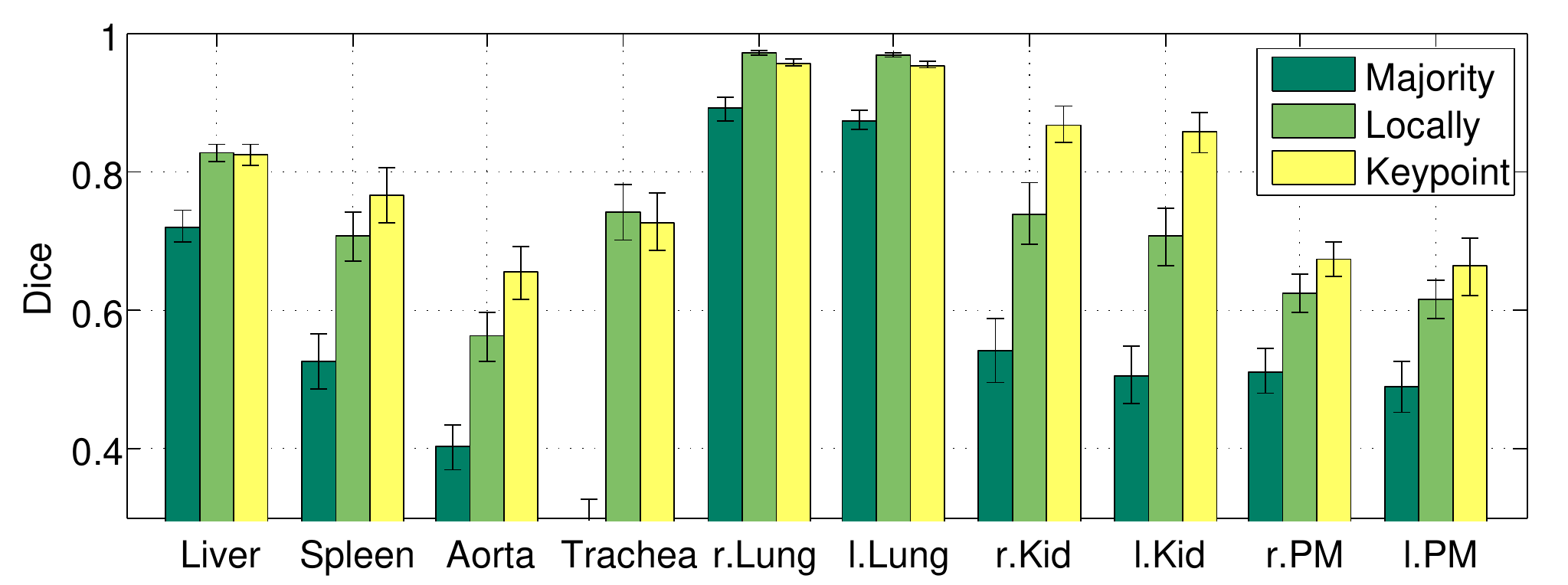}
        \caption{ceCT}
        \label{fig:gull}
    \end{subfigure}
    \begin{subfigure}[b]{0.49\textwidth}
        \includegraphics[width=\textwidth]{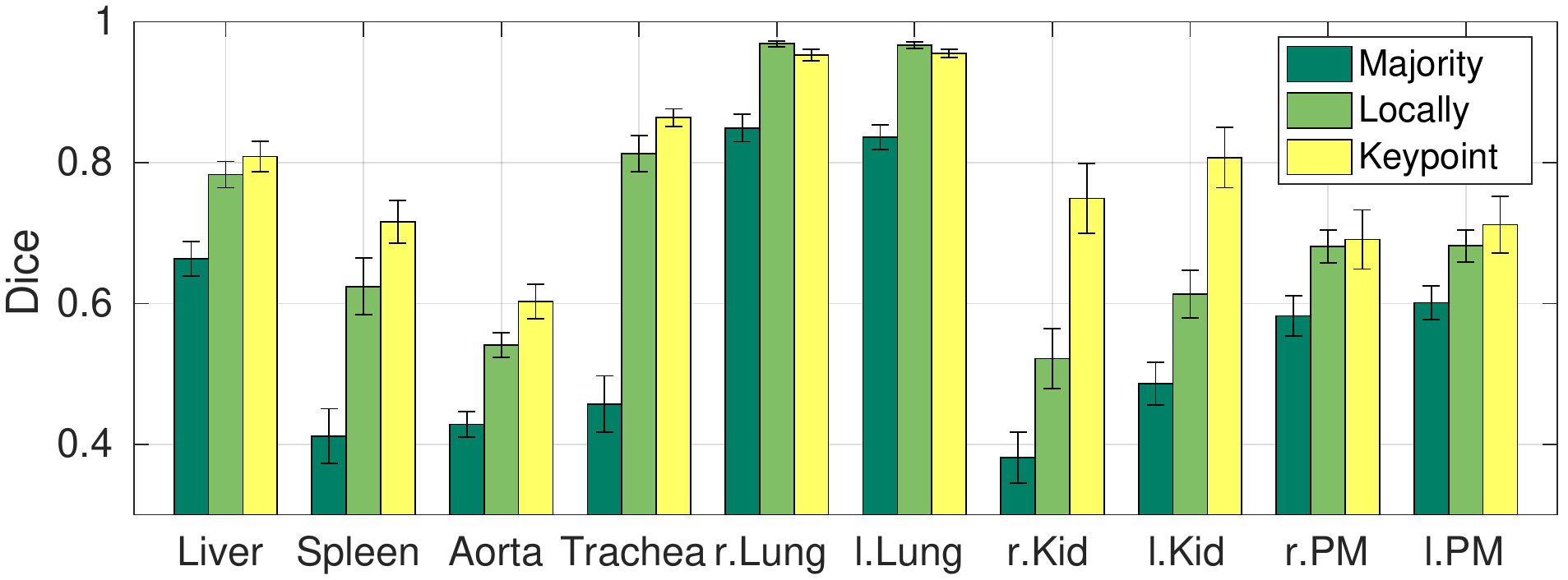}
        \caption{wbCT}
        \label{fig:gull}
    \end{subfigure}    
    \begin{subfigure}[b]{0.49\textwidth}
        \includegraphics[width=\textwidth]{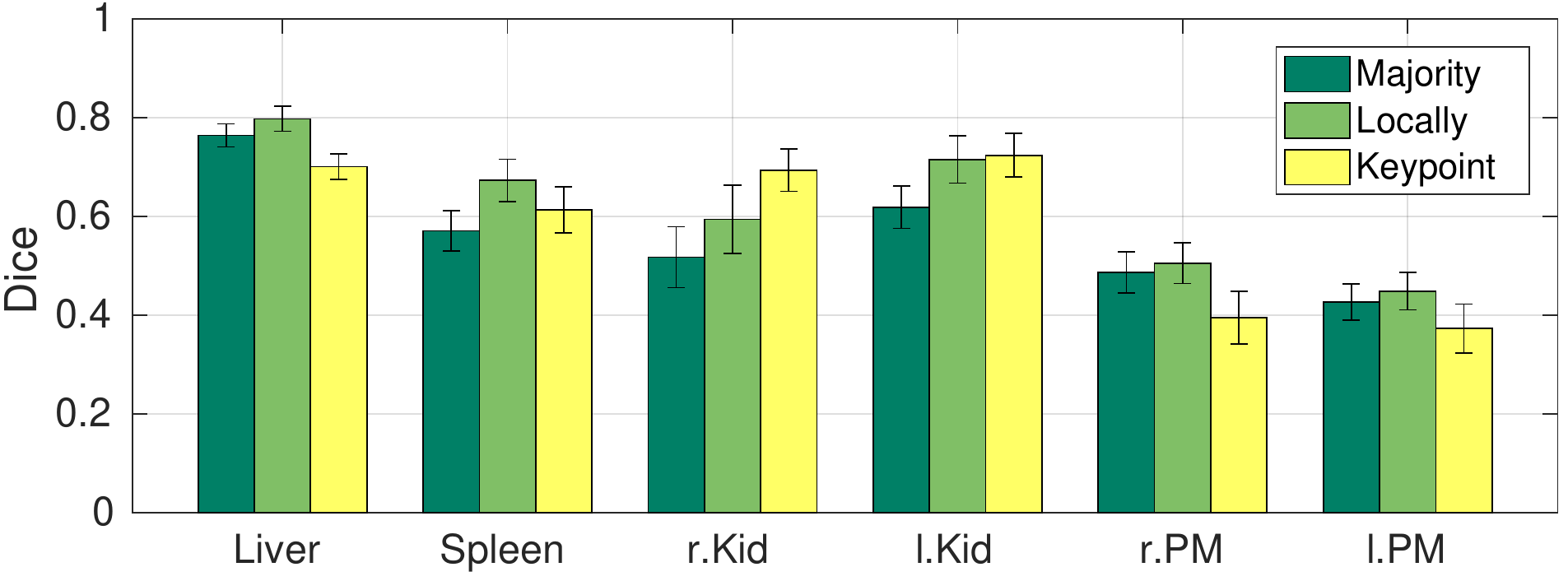}
        \caption{ceMR}
        \label{fig:gull}
    \end{subfigure}    
    \begin{subfigure}[b]{0.49\textwidth}
        \includegraphics[width=\textwidth]{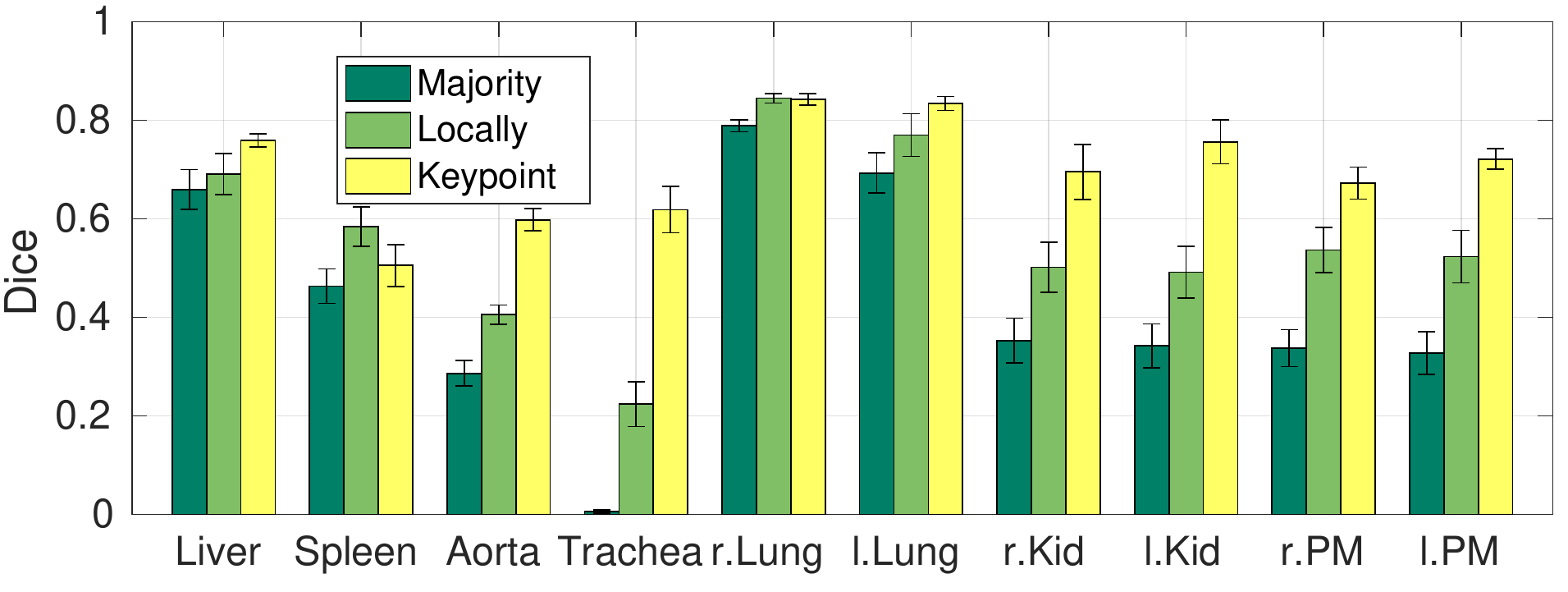}
        \caption{wbMR}
        \label{fig:gull}
    \end{subfigure}            
\caption{Segmentation accuracy on the gold corpus for different organs on ceCT, wbCT, ceMR, and wbMR images for majority voting, locally-weighted voting, and keypoint transfer. Bars indicate the mean Dice and error bars correspond to standard error.    \label{fig:ctceRes}}
\end{center}
\end{figure*}

\begin{figure}
\begin{center}
\includegraphics[width=0.35\textwidth]{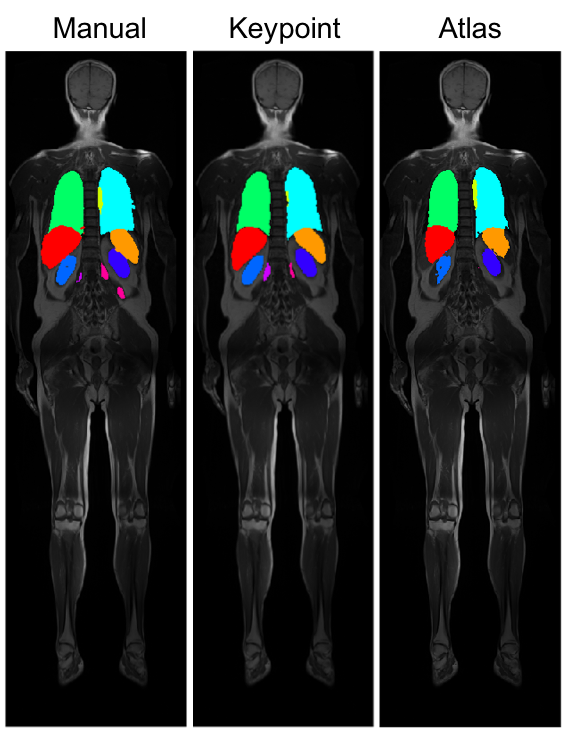} 
\caption{Example segmentation results on coronal views for wbMR overlaid on the intensity images, shown for manual, keypoint transfer, locally-weighted multi-atlas.   \label{fig:qualitative2}}
\end{center}
\end{figure}

\begin{figure}
\begin{center}
\includegraphics[width=0.5\textwidth]{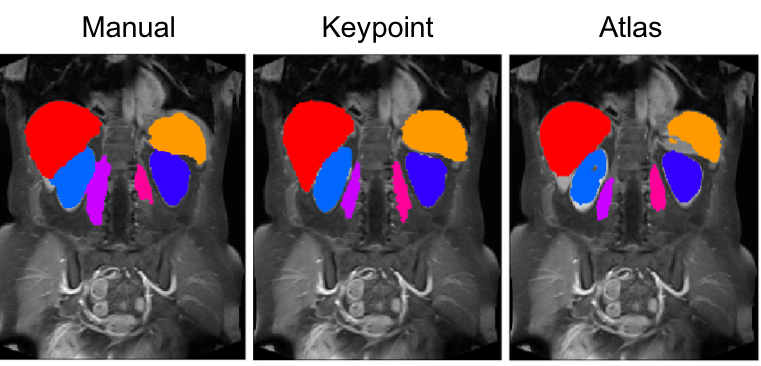} 
\caption{Example segmentation results on coronal views for ceMR overlaid on the intensity images, shown for manual, keypoint transfer, locally-weighted multi-atlas.   \label{fig:qualitative3}}
\end{center}
\end{figure}

\begin{figure*}
\begin{center}
\includegraphics[width=1.0\textwidth]{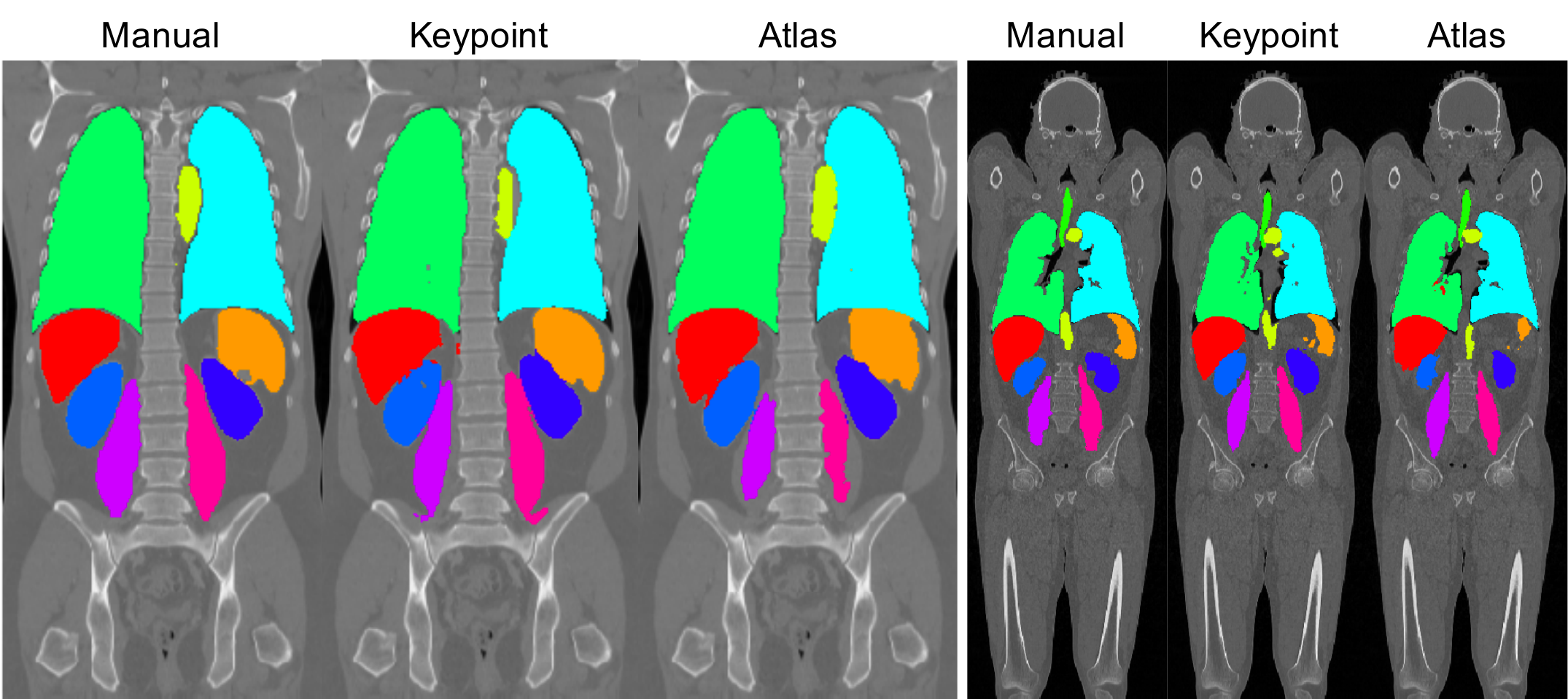} 
\caption{Example segmentation results on coronal views for ceCT (left) and wbCT (right) overlaid on the intensity images.  
Each series shows segmentations in the following order:  manual, keypoint transfer, locally-weighted multi-atlas.   \label{fig:qualitative}}
\end{center}
\end{figure*}

\section{Results}
We segment 10 anatomical structures (liver, spleen, aorta, trachea, left/right lung, left/right kidney, left/right psoas major muscle~(PM)) and 
 perform experiments on four different contrasts from the Visceral dataset re-sampled to 2mm isotropic voxels~\cite{langs2013visceral,jimenez2016cloud}: contrast-enhanced CT~(ceCT), whole-body CT~(wbCT), contrast-enhanced MR (ceMR), and whole-body MR (wbMR).  
The dataset contains 20 images for each of the contrasts with manual annotations that we refer to as gold corpus. 
Image dimensions are roughly $217 \times 217 \times 695$ for wbCT, $200 \times 200 \times 349$ for ceCT, $252 \times 87 \times 942$ for wbMR, and $195 \times 108 \times 240$ for ceMR.
All of the 10 structures are annotated on ceCT and wbCT scans, and most on wbMR. On ceMR, only liver, spleen, aorta, kidneys, and PM have sufficient annotations to allow for an evaluation. 
In addition to the gold corpus, a silver corpus exists with 65 ceCT, 62 wbCT, 71 ceMR, and 37 wbMR scans. 
The silver corpus does not have manual annotations but labels were created by fusing the results of several segmentation methods that were submitted to the Visceral challenge~\cite{jimenez2016cloud}.
The fusion was done with the SIMPLE approach~\cite{langerak2010label} and resulted in more accurate segmentations than any of the individual algorithms, but may not be as accurate as manual annotations~\cite{jimenez2016cloud}. 
\revision{Following~\cite{toews2013efficient}, we set $\sigma = 1.6$ voxels and $\kappa =\sqrt[3]{2}$ for the keypoint localization. We use 10 bins along each dimension in the Hough transform and for the distribution $p(m)$, where kernels with sigma 0.2 are employed.}

On the gold corpus, we perform leave-one-out experiments, using 19 images for training and one image for testing. 
On the silver corpus, we use the 20 images from the gold corpus as training set and the fused segmentations as reference. 
We set~$\nu=50$ for all organs, except for lungs and trachea on the CT scans, where we set~$\nu=300$. 
We compare our method to multi-atlas segmentation with majority voting~(MV)~\cite{heckemann2006automatic,rohlfing2004evaluation} and locally-weighted label fusion~(LW)~\cite{sabuncu2010generative} using ANTS~\cite{avants2008symmetric} for deformable registration\footnote{\revision{Command: ANTS 3 -m CC[.,.,1,2] -r Gauss[3,0] -t Syn[0.25]}}.  
We quantify the segmentation accuracy with the Dice volume overlap between reference and automatic segmentation.

\subsection{Gold Corpus}

Statistics for the voting on keypoint labels are displayed in  Table~\ref{tab:voting}. 
As expected, the average number of keypoints varies across organs, also influenced by an organ's size. 
Keypoints that do not receive reliable matches due to spatial constraint and distance ratio test are not labeled. 
%
Since  it is possible that certain keypoints in the test image do not appear in the training set, the focus on reliable keypoints improves the performance of the algorithm.
We observe a high voting accuracy for keypoints that are labeled.
Exceptions with lower accuracies are aorta and psoas major muscles on MR. 
Since we do not include background keypoints in the training set, all of the votes on background keypoints in the test image are incorrect. 
However, only about one third of the CT background keypoints and about one half of MR keypoints received labels. 
As long as there is no bias in transferring label maps to a specific location, the remaining background keypoints have limited impact on the segmentation outcome.

\begin{figure}
\begin{center}
\includegraphics[width=0.5\textwidth]{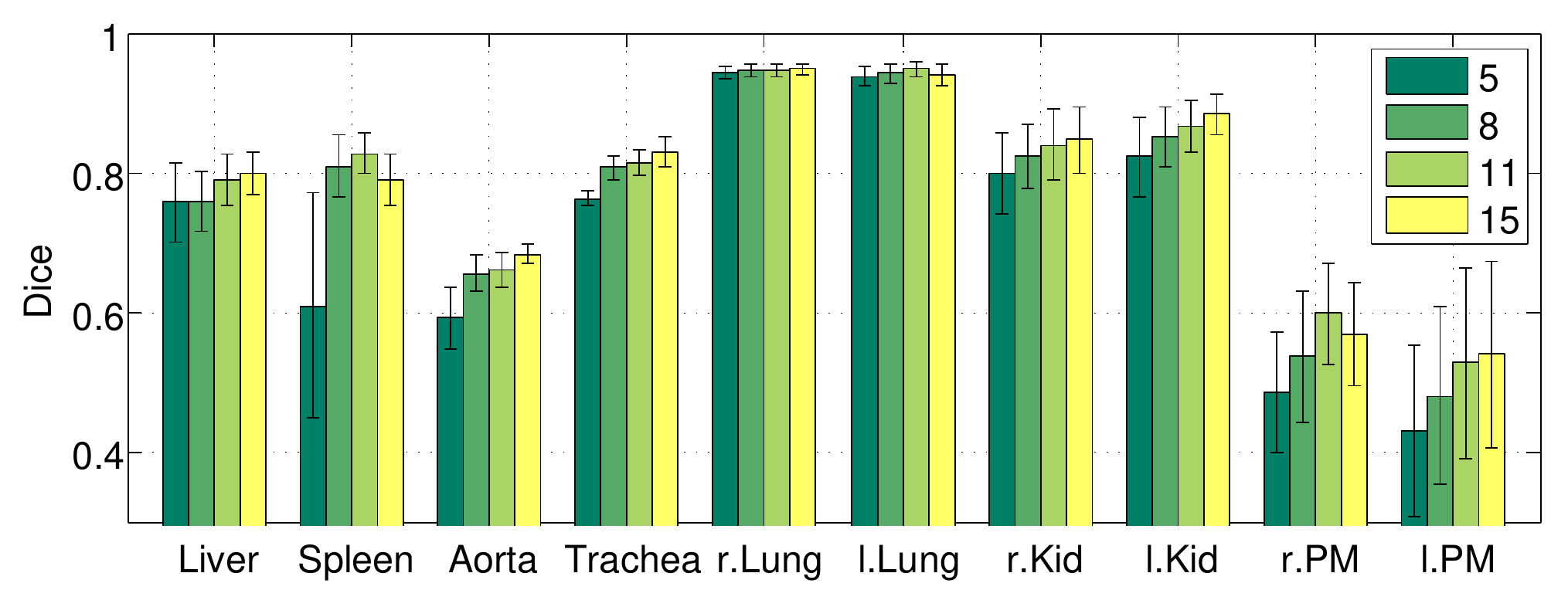} 
\caption{Segmentation results for varying the number of training images from 5 to 15. Results are shown for ten organs on ceCT images with keypoint transfer. Bars indicate the mean Dice over five test images and error bars correspond to standard error.    \label{fig:variation}}
\end{center}
\end{figure}

Fig.~\ref{fig:ctceRes} reports segmentation results for all contrasts with keypoint transfer and multi-atlas segmentation. 
Across all anatomical structures, locally-weighted voting outperforms majority voting. 
Keypoint transfer segmentation leads to segmentation accuracy comparable to that of locally-weighted voting for most structures.
\revision{Statistically significantly higher accuracy (two-tailed, paired t-test) was achieved for kidneys in ceCT (about 15 Dice points, $p<0.02$), wbCT (about 20 Dice points, $p<0.001$), and wbMR (about 20 Dice points, $p<0.005$). 
Further significant improvements exist for aorta in wbCT ($p <0.05$) and wbMR ($p<0.001$), spleen in wbCT ($p < 0.05$), trachea in wbMR ($p<0.001$), and  psoas major muscles in wbMR ($p<0.01$). 
Keypoint transfer is significantly worse than locally-weighted voting for liver in ceMR ($p<0.005$) and lungs in ceCT ($p<0.005$).} 
%
%
The transfer of label maps that are different from the keypoint label did not yield a robust improvement in these experiments. 
Figs.~\ref{fig:qualitative2} to~\ref{fig:qualitative} illustrate segmentation results for all the contrasts. 


\begin{figure}[t]
\begin{center}
\includegraphics[width=0.5\textwidth]{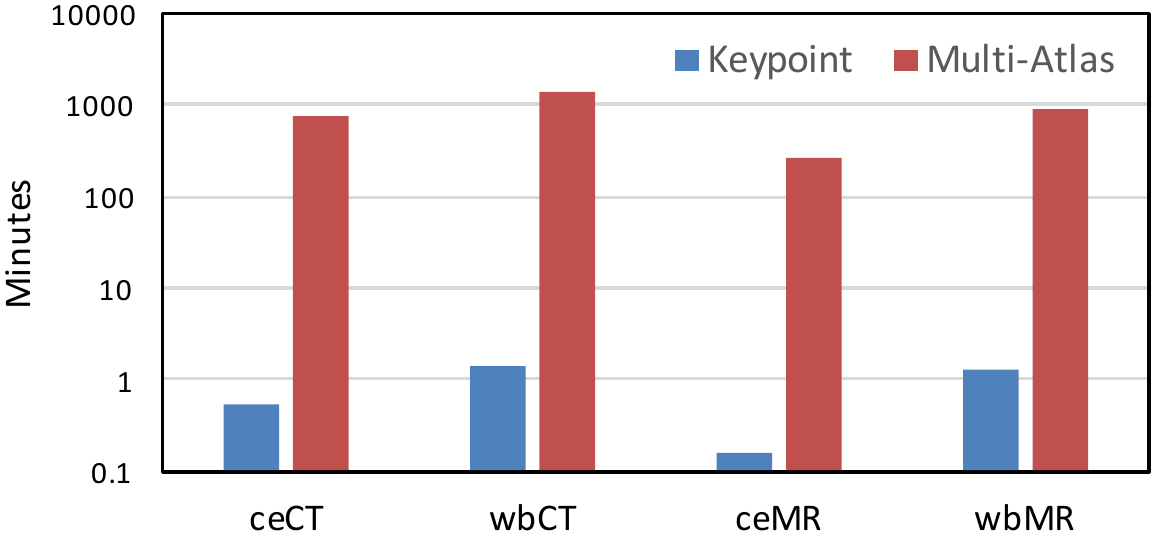} 
\caption{Average segmentation runtimes (in minutes) per image for keypoint transfer and multi-atlas label fusion with ten organs. The time is plotted on the logarithmic scale.  \label{fig:runtime}}
\end{center}
\end{figure}

\begin{figure}
\begin{center}
\includegraphics[width=0.5\textwidth]{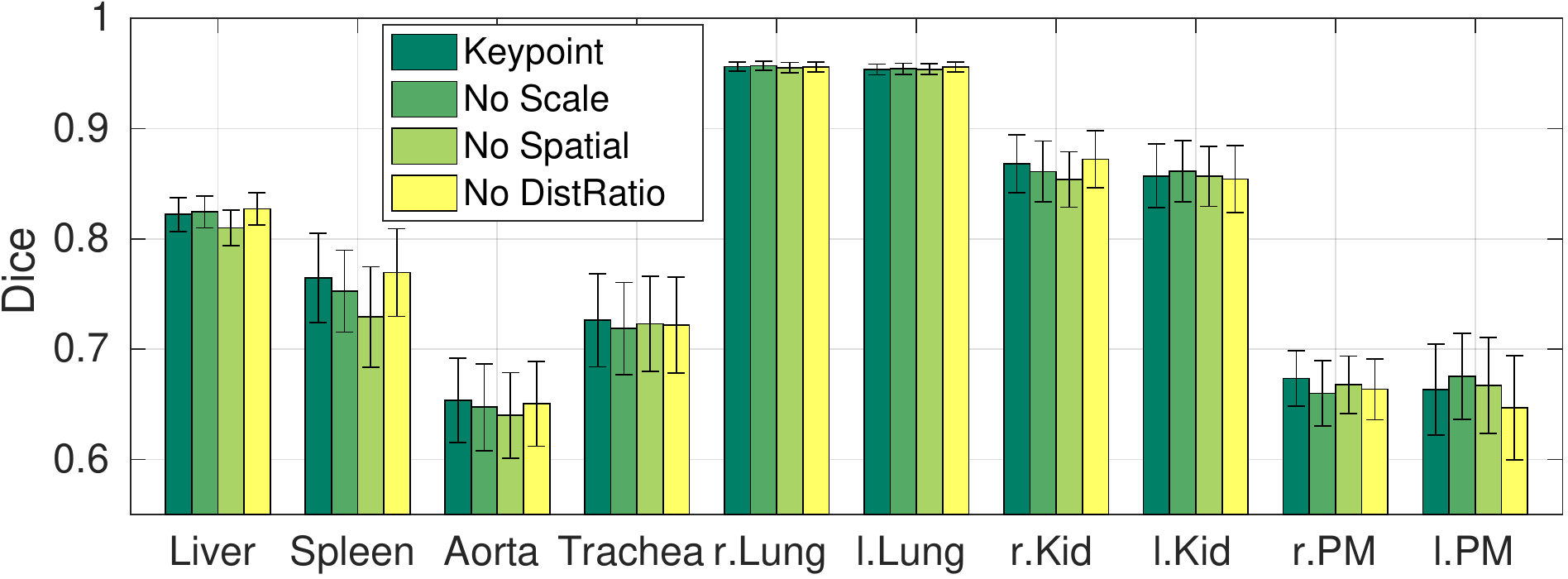} 
\caption{\revision{Segmentation results for removing the individual matching constraints. Bars indicate the mean Dice over five test images and error bars correspond to standard error. }  \label{fig:matching}}
\end{center}
\end{figure}

\begin{figure}
\begin{center}
\includegraphics[width=0.4\textwidth]{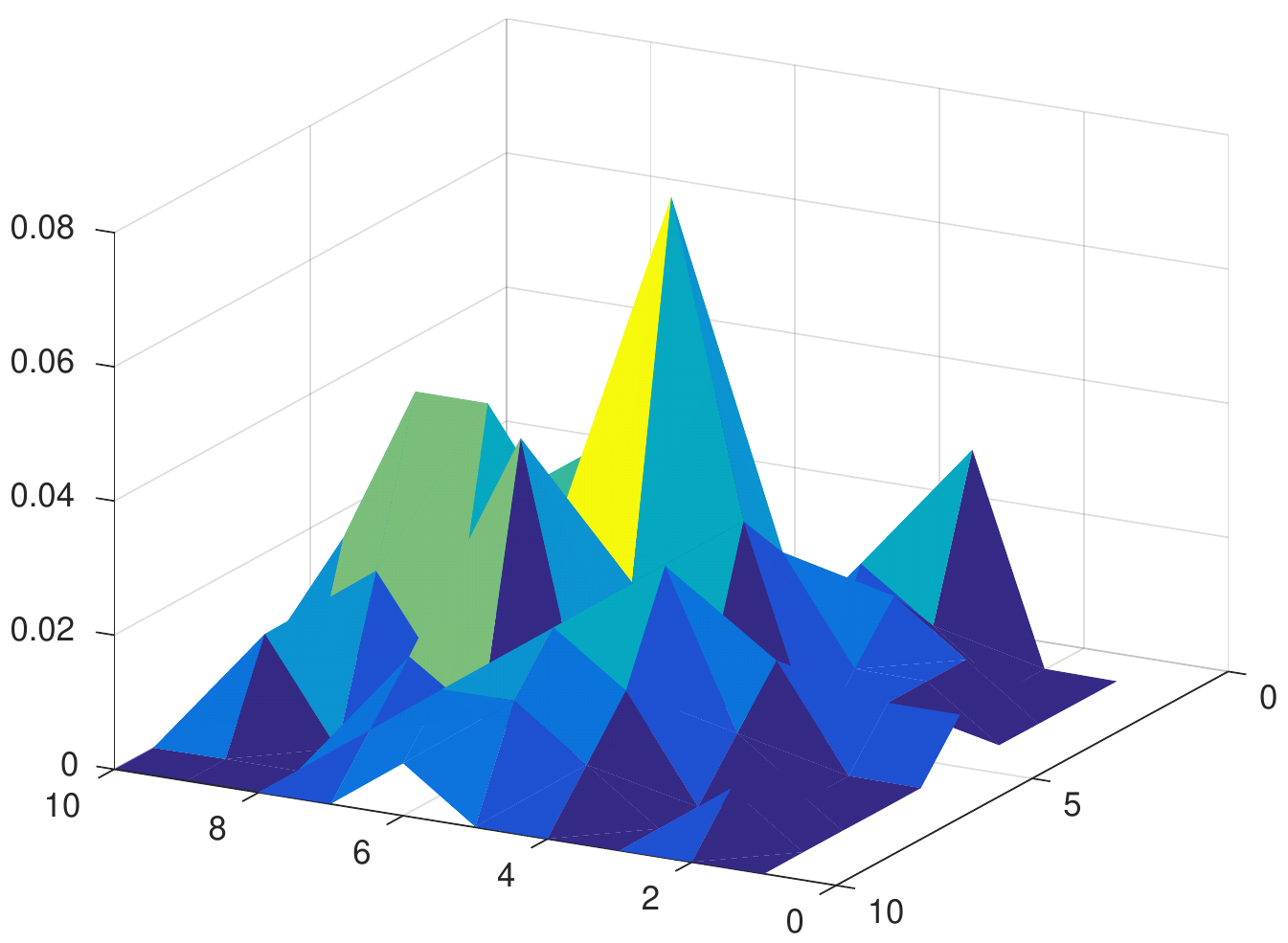} 
\caption{\revision{Visualization of the distribution over matches $p(m)$ for 2D (x-z).}  \label{fig:dist}}
\end{center}
\end{figure}

\begin{figure*}
\begin{center}
\includegraphics[width=0.7\textwidth]{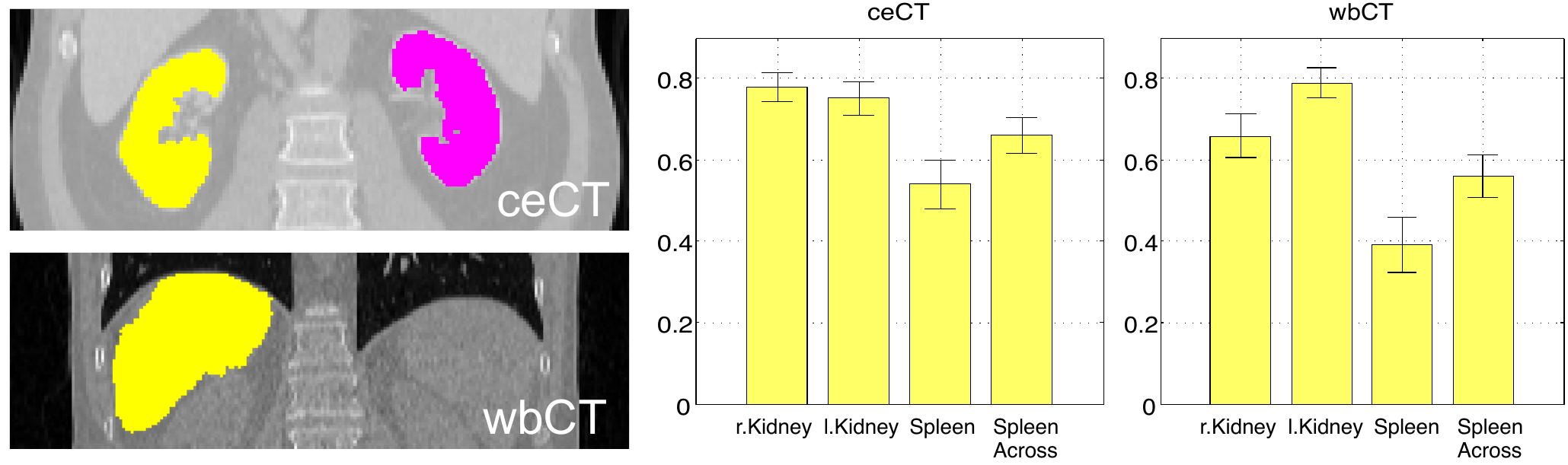} 
\caption{Segmentation results for scans with limited field-of-view. We show coronal views of images with the kidneys and the spleen for ceCT and wbCT scans. Bars indicate the mean Dice and error bars correspond to  standard error. We denote the use of lung and liver keypoints to transfer spleen segmentations as `Spleen Across'. \revision{For comparison, the mean Dice scores on the entire images for ceCT were 0.87 (r. kidney),  0.86 (l. kidney), and 0.76 (spleen), and for wbCT 0.75 (r. kidney),  0.81 (l. kidney), and 0.68 (spleen)}
\label{fig:partial}}
\end{center}
\end{figure*}

The average segmentation result for ceCT scans when varying the number of training scans from 5 to 15 is shown in Fig.~\ref{fig:variation}; the evaluation is done on the five images not included in the training set.  
As we increase the number of training images, the segmentation accuracy generally increases. 
\revision{We note a slight decrease for spleen, left lung and right PM for 15 scans, which may be due to the composition of the training set.}
Averaging over segmentations of a larger number of subjects therefore assists in recovering the true shape of the organ.
In the future, the growing number of large datasets may therefore further improve the segmentation results.
Moreover, keypoints may support the efficient implementation of an atlas selection scheme to only transfer organs from overall similar subjects.

The runtime of keypoint transfer segmentation and multi-atlas label fusion is compared in Fig.~\ref{fig:runtime}, with keypoint transfer being about three orders of magnitude faster than multi-atlas segmentation. 
For ceCT scans, as an example,  the extraction of keypoints takes about 17s and the segmentation transfer takes 16s, yielding a segmentation time for ten organs that is about half a minute. 
We implemented the segmentation transfer in Matlab without parallelization. 
The pairwise deformable registration takes most of the runtime for multi-atlas segmentation. 
To reducing computational costs for atlas-based segmentation, we also experimented with creating a probabilistic atlas. 
However, the iterative estimation of the atlas is also expensive and the high anatomical variability of the abdomen makes the summarization challenging.

\revision{We further evaluate the impact of the different matching constraints from Section \ref{sec:matching} (scale, spatial, and distance ratio) on the segmentation accuracy of ceCT data. 
We measure that 12.5\% of matches are changed due to scale, 52.6\% of matches are changed due to the distance threshold, and 60.9\% of matches are discarded due to distance ratio. 
Fig.~\ref{fig:matching} shows the segmentation accuracy for the standard keypoint matching algorithm and variations by not using scale constraints, spatial constraints, or the distance ratio test. 
Not considering spatial constraints yields to the largest decrease in accuracy. 
For scale and distance ratio, there is no unanimous picture across all organs, but the overall Dice is highest by including all of the matching constraints.  
}

\revision{We plot the distribution over matches $p(m)$ projected on the x-z axes in Fig.~\ref{fig:dist}. We observe that translations proposed by the matches do not follow a uni-modal distribution.}

\subsection*{Limited field-of-view scans}
Next to the segmentation of abdominal and whole-body scans, the segmentation of scans with limited field-of-view was also evaluated. 
In clinical practice, such partial scans are frequently acquired because of a specific diagnostic focus. 
For evaluating the performance of the algorithm, we cropped ceCT and wbCT images around the kidneys and the spleen, as shown in Fig.~\ref{fig:partial}. 
\revision{Specific for the segmentation of the spleen in the limited field-of-view scans, we noted a substantial improvement by transferring organ segmentations that are different from the keypoint label for spleen images. In this case, we also let lung and liver keypoints transfer the segmentation of spleen.}
%
Fig.~\ref{fig:partial} shows segmentation results for kidneys and spleen. 
In comparison to segmenting the whole scans, we observe a slight decrease in segmentation accuracy for partial scans. 
Overall, however, the keypoint transfer is robust to variations in the field-of-view and enables segmentation without modifications of the algorithm. 
We do not show results for the multi-atlas segmentation in this experiment because the registration between the cropped images and the training images failed. 
Since the initial alignment does not yield a rough overlap of the target regions, it is a very challenging registration problem. 
While it may be possible to design initialization techniques that improve the alignment, we view it as a major advantage of the keypoint transfer that no modification is required to handle limited field-of-view scans.

\begin{figure*}[t]
\begin{center}
    \begin{subfigure}[b]{0.49\textwidth}
        \includegraphics[width=\textwidth]{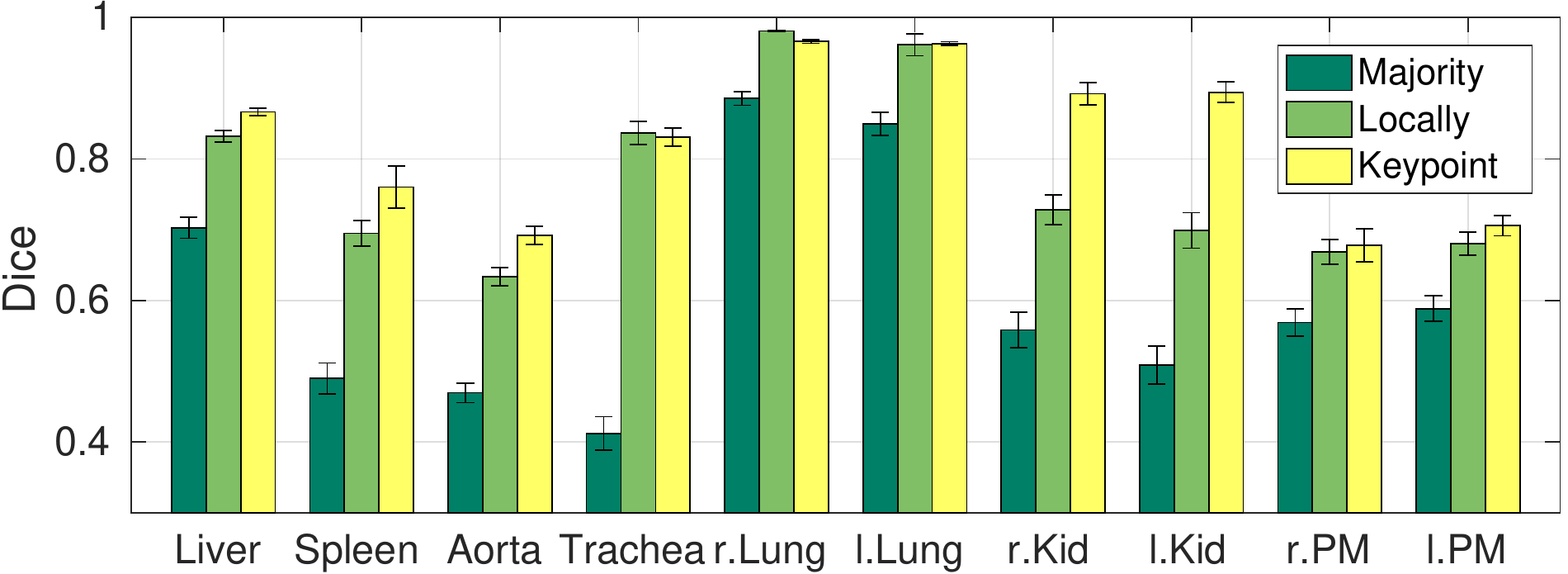}
        \caption{ceCT}
        \label{fig:gull}
    \end{subfigure}
    \begin{subfigure}[b]{0.49\textwidth}
        \includegraphics[width=\textwidth]{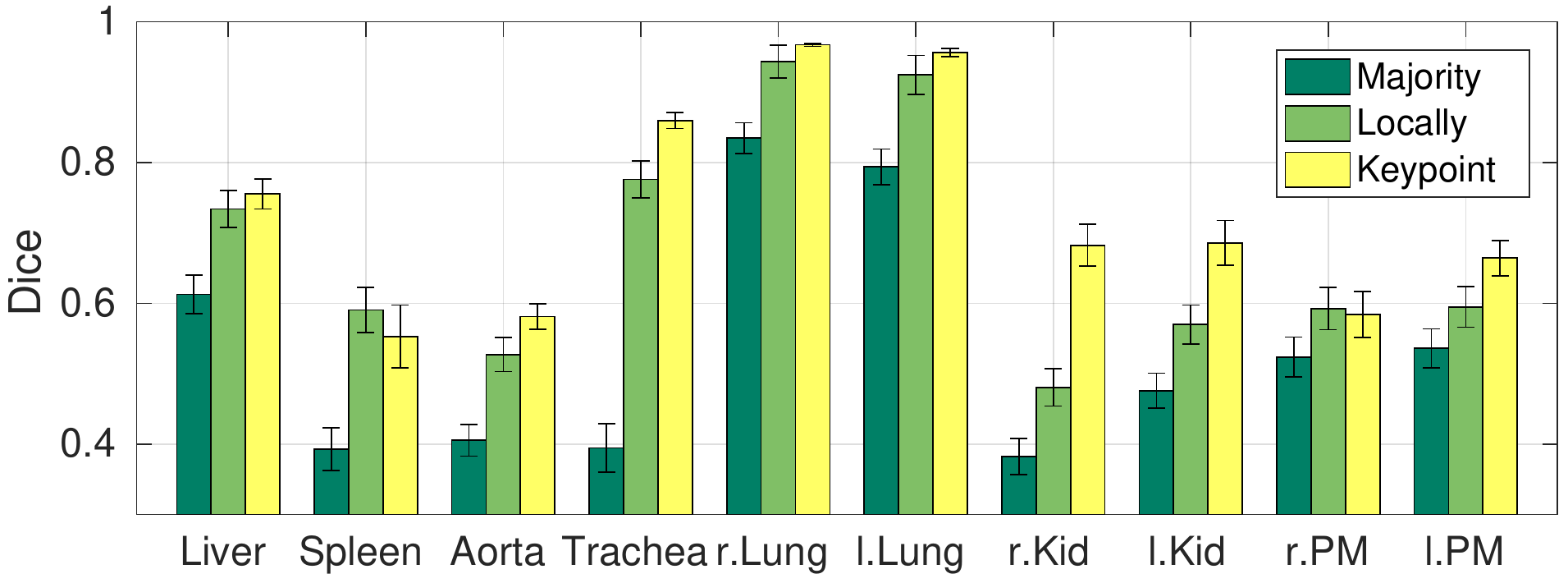}
        \caption{wbCT}
        \label{fig:gull}
    \end{subfigure}    
    \begin{subfigure}[b]{0.49\textwidth}
        \includegraphics[width=\textwidth]{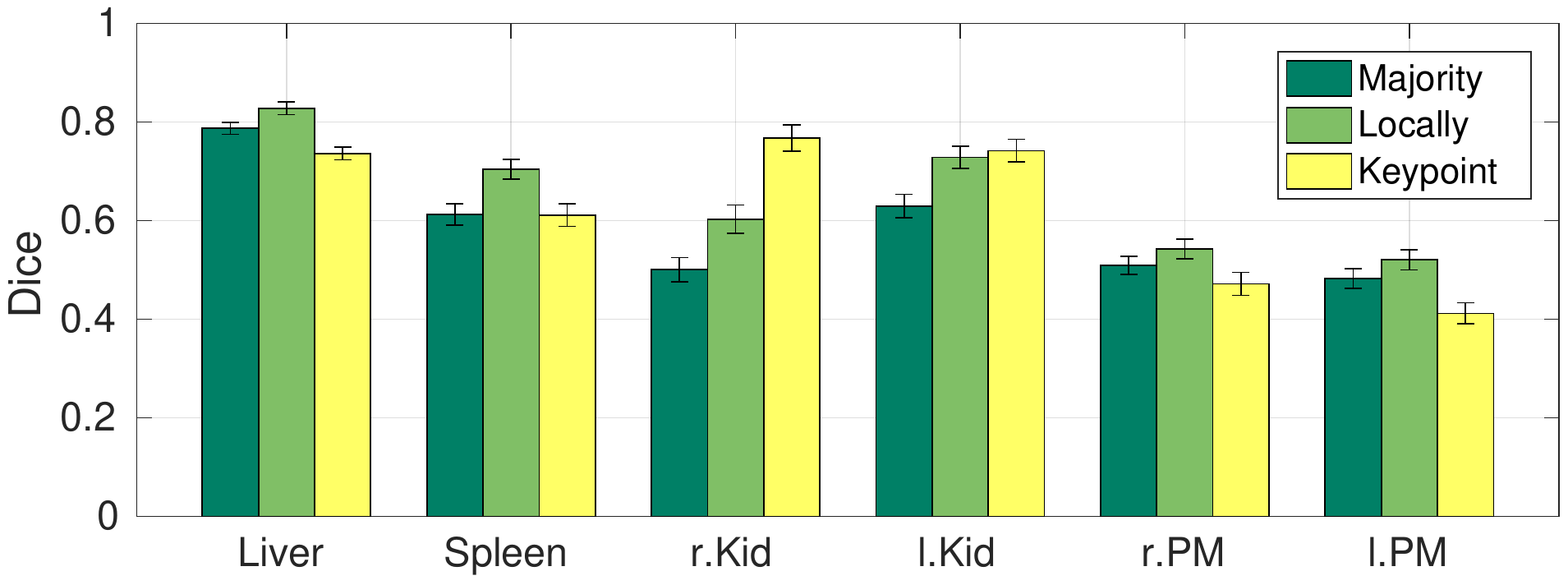}
        \caption{ceMR}
        \label{fig:gull}
    \end{subfigure}    
    \begin{subfigure}[b]{0.49\textwidth}
        \includegraphics[width=\textwidth]{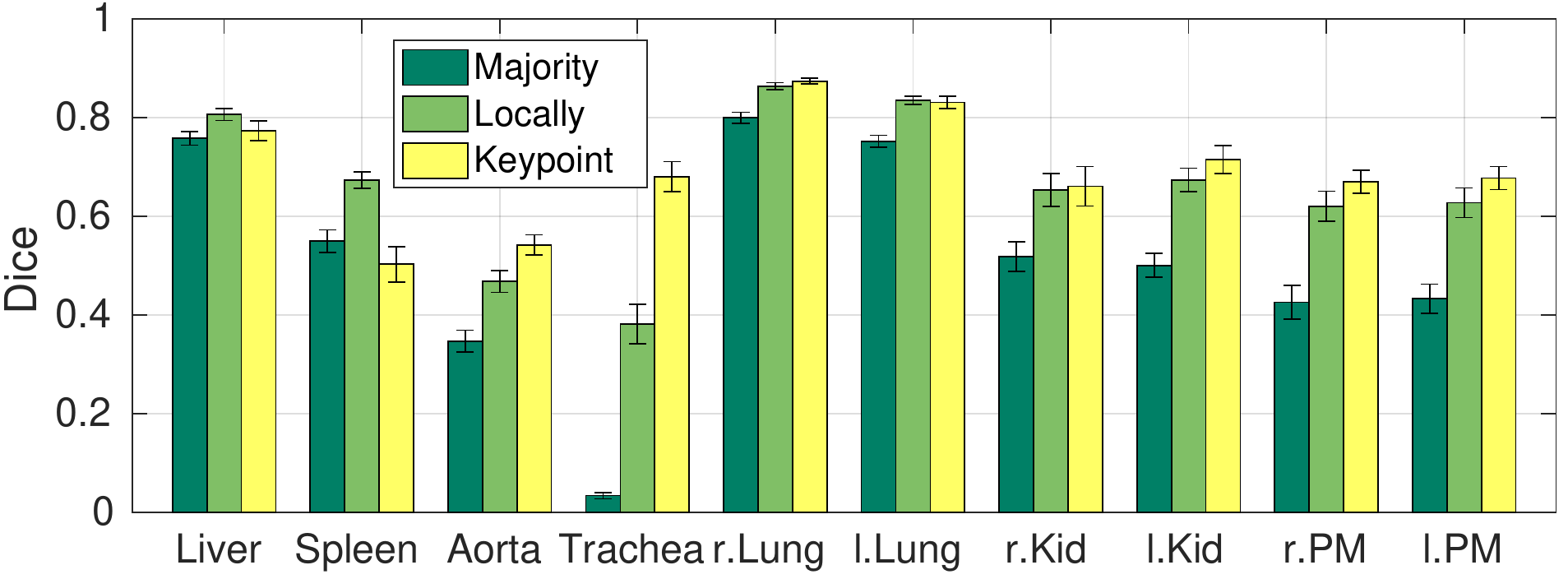}
        \caption{wbMR}
        \label{fig:gull}
    \end{subfigure}            
\caption{Segmentation accuracy on the silver corpus for different organs on ceCT, wbCT, ceMR, and wbMR images for majority voting, locally-weighted voting, and keypoint transfer. Bars indicate the mean Dice and error bars correspond to standard error.    \label{fig:silverRes}}
\end{center}
\end{figure*}

\subsection{Silver Corpus}
Fig.~\ref{fig:silverRes} reports segmentation results on the silver corpus for all contrasts, comparing keypoint transfer to multi-atlas segmentation. 
As for the gold corpus, locally-weighted voting outperforms majority voting for all anatomical structures. 
Further, we see similar results in the comparison of keypoint transfer and locally-weighted voting for the silver corpus \revision{and} for the gold corpus. 
Keypoint transfer achieves higher accuracy on kidneys for both CT contrasts. 
Liver, spleen and aorta are segmented more accurately on ceCT; aorta and trachea have higher Dice overlap for both whole-body modalities. 
Liver and spleen show lower accuracy for keypoint transfer than locally-weighted voting for both MR contrasts.


\section{Discussion}
Overall, we note a higher segmentation accuracy for CT than for MR scans. 
We think that the standardization with Hounsfield units in CT scans  is potentially beneficial at several stages of the method. 
The construction of matches as well as the keypoint voting are influenced by the descriptor similarity across scans. 
Further, the weights $W(x)$ in the keypoint segmentation are based on image intensity differences between training and test images. 
Studying the voting statistics in Table 1, we see that the voting accuracy for MR scans is comparable to CT scans for most organs. 
These good results for MR support the robustness of the descriptor to intensity variations, since the voting and matching is only based descriptor similarity. 
In contrast, the weights $W(x)$ are computed on image intensity differences between training and test images, where the variations can have a larger impact.
\revision{Approaches for correction of intensity inhomogeneity in MRI~\cite{vovk2007review} may be helpful to increase the segmentation accuracy in the future.}

In contrast to multi-atlas methods, keypoint transfer does not estimate a dense deformation field between scans. 
Instead, keypoints between organs are matched and the implied transformation is then used to transfer the organ segmentation. 
A less constrained deformation model, like this one, can have advantages in situations, where it is complicated to estimate image-wide deformation fields. 
We have shown this for the segmentation of scans with variations in the field-of-view, where keypoint transfer enabled the segmentation of partial scans. 
While we do incorporate spatial consistency of matches as weighting term~$p(m)$, this can handle multi-modal distributions and is  less restrictive than typical regularization constraints imposed on deformation fields, particularly interesting for whole-body scans with wide inter-subject variation. 

We have presented results on a gold and a silver corpus. 
The silver corpus does not have manual annotations but provides an opportunity to evaluate the segmentation methods on a larger dataset. 
Overall, the segmentation results across both corpora were similar. 
Since many of the segmentation techniques submitted to the Visceral challenge are based on multi-atlas techniques~\cite{jimenez2016cloud}, there may be a bias in the fused segmentation of the silver corpus, favoring multi-atlas methods in the comparison. 

\revision{The keypoint transfer segmentation relies on the identification of keypoints in the organs. While we did not experience this problem in our experiments, it may be a limitation for small organs without salient texture. An alternative in such scenarios could be to use keypoints from neighboring organs to transfer the segmentation. 
As shown for the spleen on the limited field-of-view scans, the transfer from neighboring organs can improve the segmentation result. 
However, we also performed experiments with using neighboring organs to transfer the label map on whole-body images and it did not lead to an improvement. A potential reason is that some organs are quite large so that keypoints can be fairly distant from the target organ. It may be a promising research direction in the future to restrict the transfer to keypoints that are close by.}

Keypoint transfer is a very fast segmentation approach. 
Depending on the image type, the segmentation of a single scans takes between 10 and 84 seconds. 
This is orders of magnitude faster than multi-atlas techniques. 
In addition, there is no training stage in the algorithm that may require additional time. 
The only preparation for a new scan to be included in the training dataset is the extraction of keypoints. 
Keypoint transfer segmentation is therefore a highly scalable approach for large training and test sets. 

 \section{Conclusion}
We proposed an approach for image segmentation with keypoints that transfers label maps of entire organs. 
The algorithm relies on sparse correspondences between keypoints in the test and training images, which increases the efficiency of the method. 
We have further demonstrated that keypoint matches are robust to variations in the field-of-view, which allowed for the  segmentation of partial scans. 
We evaluated the method on two CT and MR contrasts from gold and silver corpora. 
The accuracy of keypoint transfer segmentation compares favorably to multi-atlas segmentation, while being about three orders of magnitude faster. 
\revision{Since a segmentation can be obtained very quickly with keypoint transfer, the produced segmentation may be used as additional input for other segmentation approaches, e.g., based on deep learning.}

\textbf{Acknowledgements:} 
This work was supported in part by the Bavarian State Ministry of Education, Science and the Arts in the framework of the Centre Digitisation.Bavaria (ZD.B), the Humboldt foundation, the National Alliance for Medical Image Computing (U54-EB005149), the NeuroImaging Analysis Center (P41-EB015902), the National Center for Image Guided Therapy (P41-EB015898), and the Wistron Corporation.

\bibliographystyle{IEEEtran}

\bibliography{jab_bib}

\end{document}